\documentclass[final]{IEEEtran}

\usepackage[noadjust]{cite}
\usepackage{times}
\usepackage{epsfig}
\usepackage{graphicx}
\usepackage{amsmath}
\usepackage{amssymb}
\usepackage{enumitem}
\usepackage{subfigure}
\usepackage{booktabs}
\usepackage{multirow}
\usepackage{comment}
\usepackage{pifont}% http://ctan.org/pkg/pifont
\usepackage{siunitx}

\newcommand{\cmark}{\ding{51}}%
\newcommand{\xmark}{\ding{55}}%

\usepackage[pagebackref=false,breaklinks=true,colorlinks,bookmarks=false]{hyperref}

\begin{document}

%%%%%%%%% TITLE
\title{EarthMapper: A Tool Box for the Semantic Segmentation of Remote Sensing Imagery} 

\author{Ronald Kemker\textsuperscript{*}\thanks{\textsuperscript{*} Kemker and Gewali contributed equally to this paper.}, Utsav B. Gewali\textsuperscript{*}, \& Christopher Kanan,~\IEEEmembership{Member,~IEEE}% <-this % stops a space

\thanks{R. Kemker, U. Gewali, and C. Kanan are with the Carlson Center for Imaging Science, Rochester Institute of Technology, Rochester, NY, 14623 USA.}% <-this % stops a space

}

\maketitle
%\thispagestyle{empty}

%%%%%%%%% ABSTRACT
\begin{abstract}
Deep learning continues to push state-of-the-art performance for the semantic segmentation of color (i.e., RGB) imagery; however, the lack of annotated data for many remote sensing sensors (i.e. hyperspectral imagery (HSI)) prevents researchers from taking advantage of this recent success.  Since generating sensor specific datasets is time intensive and cost prohibitive, remote sensing researchers have embraced deep unsupervised feature extraction.  Although these methods have pushed state-of-the-art performance on current HSI benchmarks, many of these tools are not readily accessible to many researchers.  In this letter, we introduce a software pipeline, which we call EarthMapper, for the semantic segmentation of non-RGB remote sensing imagery.  It includes self-taught spatial-spectral feature extraction, various standard and deep learning classifiers, and undirected graphical models for post-processing.  We evaluated EarthMapper on the Indian Pines and Pavia University datasets and have released this code for public use.
\end{abstract}

\begin{IEEEkeywords}
Hyperspectral imaging, self-taught learning, feature learning, deep learning, autoencoder, graphical model.
\end{IEEEkeywords}

\IEEEpeerreviewmaketitle

%%%%%%%%% BODY TEXT
\section{Introduction}

\IEEEPARstart{S}{emantic} segmentation is automatically labeling every pixel in an image with a semantic category, which is used by the remote sensing community for land-cover classification~\cite{kussul2017deep}, vegetation quality estimation~\cite{tong2018deep}, among others. In computer vision, results on semantic segmentation benchmarks~\cite{long2015fully,pascal-voc-2012} have rapidly increased thanks to deep convolutional neural networks (DCNNs) that are pre-trained on millions of annotated RGB images (e.g., from ImageNet).  While there is a large amount of annotated, publicly available RGB imagery for training large DCNNs, annotated data for specialized imaging systems, such as multispectral and hyperspectral imagery (HSI), is scarce because generating large, sensor specific datasets is expensive and man-power intensive. To compensate, the remote sensing community has embraced unsupervised~\cite{kemker2017self} and semi-supervised methods~\cite{kemker2018low,zhan2018semisupervised} for the semantic segmentation of overhead HSI.  

\begin{figure}[ht!]
    \centering
    \includegraphics[width=0.65\linewidth]{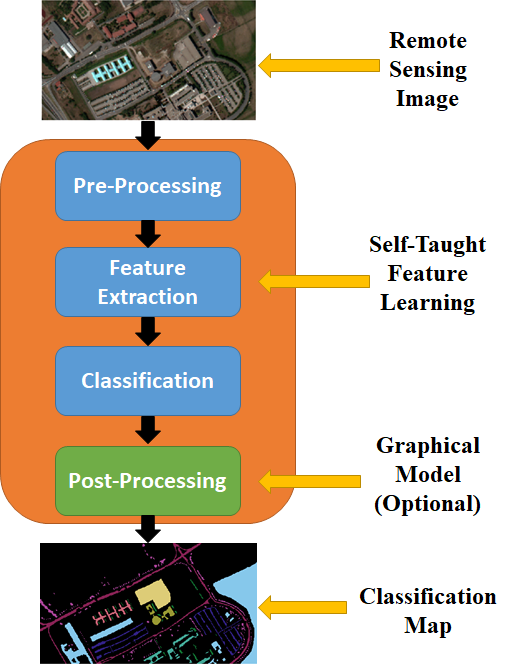}
    \caption{Our EarthMapper's pipeline for the semantic segmentation of remote sensing imagery.}
    \label{fig:pipeline}
\end{figure}

Deep unsupervised methods have shown some success at HSI semantic segmentation~\cite{lin2013spectral, tao2015unsupervised}. These methods extract spatial-spectral features from an HSI using a deep network that is typically trained to reconstruct its input (subject to some constraints), which results in it learning how to extract good features for this reconstruction task. The most common approach is to use a stacked convolutional autoencoder. After training, the encoder portion of the network is used to extract features that are then passed to some sort of classifier (e.g., a support vector machine (SVM)).  The majority of past work has trained their deep feature extraction networks directly on the image that will be classified. This is  computationally expensive and the features learned may not generalize to other datasets. Instead, in self-taught feature learning,  spatial-spectral features are directly learned from a large corpus of unlabeled HSI~\cite{kemker2017self}.  Self-taught learning methods acquire a more generalized set of spatial-spectral features, enabling state-of-the-art segmentation performance across multiple datasets without retraining the feature extraction model per dataset. 

Current software tool boxes for remote sensing classification do not include deep learning methods for feature extraction or classification~\cite{Grizonnet2017,spectral}.  In this letter, we introduce  EarthMapper, a semantic segmentation pipeline for non-RGB remote sensing data that uses self-taught learning.  EarthMapper includes undirected graphical models (UGMs), which are used as a post-processing technique to improve classification performance. One of the included UGMs is a fully-connected conditional random field, which has not been previously evaluated for HSI segmentation.

We demonstrate EarthMapper's utility on two standard HSI benchmarks: Indian Pines and Pavia University. EarthMapper has been publicly released on GitHub.  It is modular, so it can be easily modified to incorporate future advancements.

\section{Related Work}
Unsupervised feature extraction has been remarkably successful for sensors that have little annotated data available and currently represents the state-of-the-art solution for classifying HSI.  Researchers have developed a number of  spatial-spectral feature extraction methods that learn the features directly from the data~\cite{liu2015hyperspectral,lin2013spectral,zhao2015combining,ma2015hyperspectral}.  Because they are learned from a single HSI, these features may perform poorly with other datasets. Learning features per image is computationally expensive for processing large quantities of HSI~\cite{tao2015unsupervised}.  Self-taught feature learning was recently introduced as a method to build spatial-spectral feature extracting frameworks that generalize well across a variety of different scenes and sensors~\cite{kemker2017self}.  These frameworks 1) learn their feature representation from a large quantity of unlabeled HSI, 2) are used to extract discriminative features from the smaller labeled datasets, and 3) the features are fed to a standard classifier (e.g., SVM).

UGMs have been widely used in HSI classification to produce smooth, coherent segmentation maps by enforcing spatial contexts~\cite{gewali2018tutorial}. Their main advantage is that they can be easily combined with other classification approaches as a post-processing method that improves accuracy. UGMs have been used with classifiers trained directly on image pixels~\cite{tarabalka2010svm} and on image derived features~\cite{xia2015spectral}. Although researchers have started to combine deep learning with UGMs for HSI~\cite{alam2016crf,cao2018hyperspectral}, earlier works focused on supervised feature learning rather than unsupervised.  Most UGM methods for HSI have employed a simple grid-structured pairwise model. While more advanced higher-order models, e.g., robust $P^n$, have been investigated for HSI~\cite{zhong2011modeling}, these models require a large number of training pixels. A better approach is to use a  fully-connected pairwise model~\cite{krahenbuhl2011efficient}, which uses a graph with edges between every node rather than only 4-connected nodes as in grid-structured models; however, until this work, they had not been tested for the semantic segmentation of HSI. 

Popular hyperspectral remote sensing libraries, such as Spectral Python and Orfeo, and applications, such as ENVI and eCognition, contain only traditional methods for HSI segmentation, and do not carry deep feature learning or graphical models.  To make these methods more easily accessible to the broad remote sensing community, we have created the EarthMapper Tool Box. 

\section{EarthMapper Tool Box}

EarthMapper is a modular framework for HSI semantic segmentation (see Fig.~\ref{fig:pipeline}), which we have released for public use\footnote{\url{https://github.com/rmkemker/EarthMapper}}.  EarthMapper has a variety of pre-trained self-taught feature learning frameworks (Sec.~\ref{section:self-taught}), which can be used to extract spatial-spectral features from annotated HSI data.  These features are passed to a classifier that outputs a preliminary classification map, which can then be cleaned-up with one of two UGM post-processing methods (Sec.~\ref{section:graph}).   

\subsection{Self-Taught Feature Learning}
\label{section:self-taught}
EarthMapper has three different pre-trained self-taught feature learning frameworks: multi-scale independent component analysis (MICA)~\cite{kemker2017self}, stacked convolutional autoencoder (SCAE)~\cite{kemker2017self}, and the stacked multi-loss convolutional autoencoder (SMCAE).  All three of these frameworks are trained on large quantities of unlabeled HSI from three different sensors: NASA's Airborne Visible/Infrared Imaging Spectrometer (AVIRIS), NASA's EO-1 Hyperion, and Goddard's LiDAR, Hyperspectral \& Thermal Imager (GLiHT).

\begin{figure*}[t]
    \centering
    \includegraphics[width=0.95\linewidth]{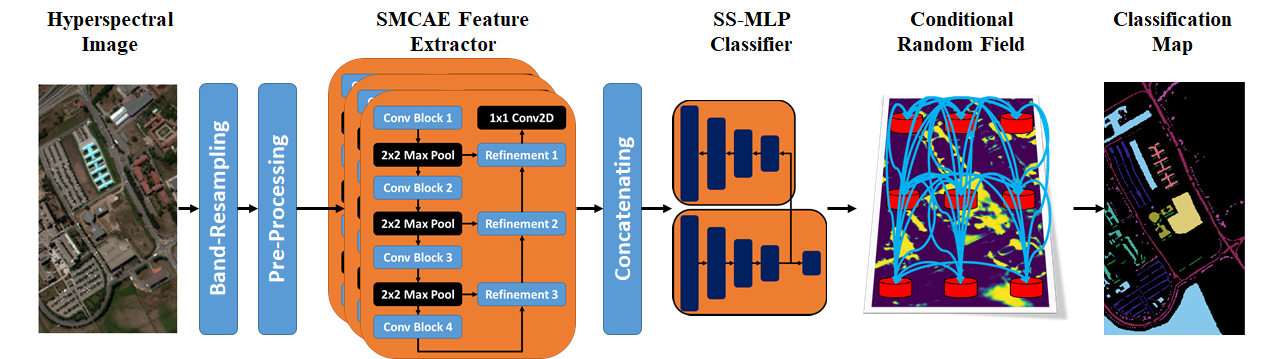}
    \caption{The EarthMapper pipeline configuration used in our experiments.  It uses SMCAE for feature extraction, SS-MLP for classification, and a fully-connected conditional random field for post-processing.\label{fig:pmica}}
\end{figure*}

MICA is a shallow feature extractor that convolves a bank of learned filters with an input HSI.  It uses ICA to learn bar/edge, gradient, and blob detecting filters from image data. The learned filters resemble receptive field properties of neurons in primary visual cortex. In \cite{kemker2017self},  MICA worked reasonably well at classifying benchmark HSI datasets, but its optimal receptive field size was dependent on the GSD of the dataset to be classified. 

SCAE~\cite{kemker2017self} is a deep feature extracting framework. While autoencoders had been used in prior work with remote sensing data,  SCAE was shown to work better on  multiple datasets because it was trained on large quantities of publicly-available remote sensing imagery rather than on a single image.  This allowed the learned features to generalize and transfer well between different datasets.  The issue with SCAE is that backpropagation during training tends to do a better job at correcting errors closer to the input layers and neglects the errors in deeper layers~\cite{valpola2015neural}.  SMCAE is an improved self-taught learning framework that uses a weighted sum of the reconstruction losses between each symmetrical encoder-decoder layer in the autoencoder, which allows backpropagation to correct errors in both shallow and deeper layers~\cite{kemker2018low}.

\subsection{Undirected Graphical Models for Post-Processing}
\label{section:graph}
UGMs factorize the joint probability for a set of variables over the cliques of a undirected graph, such that each variable is conditionally independent of other variables given its neighbors~\cite{koller2009probabilistic}. They can be used to model the spatial dependencies between the labels of neighboring pixels in HSI. When applied to images, the label associated with every pixel in the image is represented by a graph node and edges represent the relationship between the labels. The expressiveness of the UGMs is controlled by the structure of the graph and energy functions defined over the graph's cliques. The most common type of UGM used for HSI classification is a pairwise model~\cite{gewali2018tutorial}. It defines the joint distribution of the pixel labels of an image as   
\begin{equation}
\footnotesize
p(\mathbf{y}) = \frac{1}{Z} \exp \left( -E\left(\mathbf{y}\right) \right),  
\end{equation}
where $\mathbf{y} = \left[y_1,...,y_N\right]^T$ is a vector containing labels of all $N$ pixels in the image, $E\left(\mathbf{y}\right)$ is the total energy and  $Z = \sum_{\mathbf{y}} \exp \left( -E\left(\mathbf{y}\right) \right)$ is the partition function. The total energy is equal to the sum of unary energies and pairwise energies defined over all nodes and edges of the graph, such that
\begin{equation}
\footnotesize
E\left(\mathbf{y}\right) = \sum_{i \in V} E_i\left(y_i\right) + \sum_{(i,j) \in D} E_{ij}\left(y_i,y_j\right),
\end{equation}
where $V$ is the set of all nodes, $D$ is the set of all edges, $E_i\left(y_i\right)$ is the unary energy function at $i^{th}$ pixel, and $E_{ij}\left(y_i,y_j\right)$ is the pairwise energy function defined at the edge between $i^{th}$ and $j^{th}$ pixels. The unary energy captures the spectral information at the pixel location while the pairwise energy captures the spatial relationships between a pair of pixels. The UGMs whose energy functions are not dependent on input features are called the Markov random fields (MRF) and the ones whose energy functions are dependent on the input features are called the conditional random fields (CRF). 

The inference about the pixel labels, $\mathbf{y}$, is generally performed by maximum a posteriori estimation, which is equivalent to minimizing the total energy, $E\left(\mathbf{y}\right)$, by an optimization algorithm such as GraphCuts~\cite{boykov2001fast}. There are usually parameters associated with the energy functions. These can be optimized using maximum likelihood estimation; however, when there are only a few parameters, they are usually tuned using grid-search over the validation set.

EarthMapper includes grid-structured and fully-connected pairwise UGMs for post-processing. The grid-structured model is the most common type used in HSI semantic segmentation, and a detailed description can be found in \cite{gewali2018tutorial}. The fully-connected model, as  described in this letter, has not been previously explored for HSI segmentation. A fully-connected pairwise model has an edge between every pair of nodes in the graph. It is more expressive than the grid-structured model as each node is not just connected to its 4-neighbors. 

The unary energy function is given by the negative logarithm of the class probability predicted by the classifier, $E_{i}\left(y_i\right) = -\log \left( P(y_i|x_i)\right)$. The pairwise energy function used depends on the spatial location of the pixels and is given by 

\begin{equation}
\footnotesize
E_{ij}\left(y_i,y_j\right) = 
\begin{cases}
    0,& \text{if } y_i = y_j\\
    w_1 \exp \left( - \frac{{\left|\mathbf{p}_i-\mathbf{p}_j\right|}^2}{2 {\theta}_{\gamma}^2} \right),              & \text{otherwise},
\end{cases}
\end{equation}
where, $\mathbf{p}_i = [r_i,c_i]$ and $\mathbf{p}_j = [r_j, c_j]$ are the spatial coordinates of $i^{th}$ and $j^{th}$ pixels with $r_i$ and $c_i$ being the row and the column numbers of the $i^{th}$ pixel and $r_j$ and $c_j$ being the row and the column numbers of the $j^{th}$ pixel. $w_1$ and ${\theta}_{\gamma}$ are scalar parameters that are tuned using the validation set.  This pairwise energy function promotes the labeling of pixels which are spatially close to one another in the same class. Since, the number of edges in a fully connected graph grows at $O(n^2)$ with the number of nodes, it is inefficient to perform inference in such models using standard algorithms, especially when the image size is large. In this case, the mean-field approximate inference can be used to efficiently make predictions~\cite{krahenbuhl2011efficient}.

\subsection{Existing HSI Analysis Tool Boxes}

There are multiple remote sensing clustering and semantic segmentation software packages available, which are compared in Table~\ref{table:toolboxes}.  The Geospatial Data Abstraction Library (GDAL)~\cite{GDAL} is a popular open-source C/C++ library used to read, write, and process remote sensing imagery.  ENVI is the leading remote sensing software package, and has several basic processing and semantic segmentation algorithms built-in, but it is not open-source.  Spectral Python is an open-source Python library that has several image processing, clustering, and classification algorithms built-in~\cite{spectral}. The Orfeo Toolbox~\cite{Grizonnet2017} and RSGISLib~\cite{Bunting:2014:RSG:2560972.2561467} are both open-source libraries that specialize in Geographic Object Base Image Analysis (GEOBIA) based classification.  They both have some advanced feature extraction methods that can be used to improve performance.  eCognition~\cite{flanders2003preliminary} and InterImage~\cite{alexandre2018interimage} are GUI based frameworks that also specialize in GEOBIA classification.  InterImage is open-source; whereas, eCognition is proprietary.  All of these packages do not include deep learning support, and many of them do not have any advanced spatial-spectral feature extraction methods built-in.   

\begin{table}[t]
\centering \footnotesize
\setlength\tabcolsep{1.5pt}
\caption{Comparison of Hyperspectral Image Analysis Tool Boxes.  The EarthMapper pipeline includes low- and deep-level feature extraction, various supervised classifiers, and undirected graphical models (UGM) for post-processing.}
\begin{tabular}{@{}r|c|c|c|c|c|c@{}}\toprule
\multirow{2}{*}{\textbf{Tool Box}} & \textbf{Open-} & \textbf{Supervised} & \multicolumn{2}{c|}{\textbf{Feature Extraction}} & \multirow{2}{*}{\textbf{UGM}}  & \textbf{Language/} \\
&\textbf{Source} & \textbf{Classification} &  \textbf{Low-Level} & \textbf{ Deep}&  & \textbf{Interface}   \\
\midrule
ENVI & \xmark & \cmark & \cmark & \xmark& \xmark & GUI/IDL \\
Orfeo Toolbox & \cmark & \cmark & \cmark & \xmark & \xmark & C/C++  \\
GDAL & \cmark & \xmark & \xmark & \xmark & \xmark & C/C++  \\
Spectral & \cmark & \cmark & \xmark & \xmark & \xmark & Python  \\
RSGISLib & \cmark & \cmark & \xmark & \xmark & \xmark & Python  \\
eCognition & \xmark & \cmark & \xmark & \xmark & \xmark & GUI  \\
InterImage & \cmark & \cmark & \xmark & \xmark & \xmark & GUI   \\
\textbf{EarthMapper} & \cmark & \cmark & \cmark & \cmark & \cmark & Python  \\
\bottomrule 
\end{tabular}
\label{table:toolboxes}
\end{table}

\section{Experiments and Results}
\subsection{Stacked Multi-Loss Convolutional Autoencoder (SMCAE)}
We use SMCAE to extract spatial-spectral features from the source image that will be classified.  SMCAE is pre-trained on open-source imagery acquired by AVIRIS, GLiHT, and Hyperion sensors.  These images are of a diverse set of scenes, and for AVIRIS, at multiple scales.  Prior to passing the data to SMCAE, we re-sample the spectral bands of the input HSI to match that of the target image (i.e., HSI used to train feature extractor).  We pass the re-sampled data through a feature extractor and then scale the output to zero-mean/unit-variance prior to passing it to the classifier.  Specific implementation details can be found in \cite{kemker2018low}.

\subsection{Semi-Supervised Multilayer Perceptron (SS-MLP)}
We use a SS-MLP neural network for classification.  It takes a feature vector extracted from either the raw HSI cube or SMCAE and assigns each pixel a set of probabilities that it belongs to a given class.  We can either 1) take the index (i.e., argmax) of the maximum probability to give us our class label or 2) pass those probabilities to the CRF to post-process the classification map, which reduces salt-and-pepper classification errors.  We cross-validate (see experiments for more details) to determine the optimal number of hidden layers (2-10), units per hidden layer (64-3000), and weight decay parameter for regularization.  It is trained with a mini-batch size of 8, and we used Nadam~\cite{dozat2016incorporating} to optimize SS-MLP.  We start with an initial learning rate of $0.002$ and drop the learning rate by a factor of 10 when the validation loss plateaus.  We stop training when the validation loss does not decrease over a period of 50 epochs.  More details can be found in \cite{kemker2018low}.

\subsection{Fully-Connected CRF}
The parameters of the post-processing CRF were tuned using grid-search over a range of $[10^\text{-3},10^3]$ and 30 iterations of mean field approximation were used for inference.  

\subsection{Data Description}
We experiment on two widely used public datasets--Indian Indian and Pavia University. They are described in Table~\ref{tab:datasets}.
\begin{table}[t]
\centering
\caption{Datasets evaluated in this letter.}
\label{tab:datasets}
\begin{tabular}{@{}l|cc@{}}
\toprule
& \textbf{Pavia University} & \textbf{Indian Pines}\\
\midrule
\textbf{Sensor} & ROSIS & AVIRIS\\
\textbf{Location} & Pavia, Italy & Northwest Indiana\\
\textbf{Scene} & Urban & Agricultural\\
\textbf{Spectral Range} &  \SIrange{430}{860}{\nano\metre} & \SIrange{400}{2500}{\nano\metre}\\
\textbf{Spatial Dimensions} & 610 $\times$ 340 & 145 $\times$ 145\\
\textbf{Ground Sample Distance} & \SI{1.3}{\meter} & \SI{20}{\meter}\\
\midrule
\multicolumn{3}{l}{ROSIS - Reflective Optics System Imaging Spectrometer}\\
\multicolumn{3}{l}{AVIRIS - Airborne Visible/Infrared Imaging Spectrometer}\\
\bottomrule
\end{tabular}
\end{table}

\begin{figure}[t!]
  \centering
  \subfigure[Pavia University]{%
  \includegraphics[height=3cm]{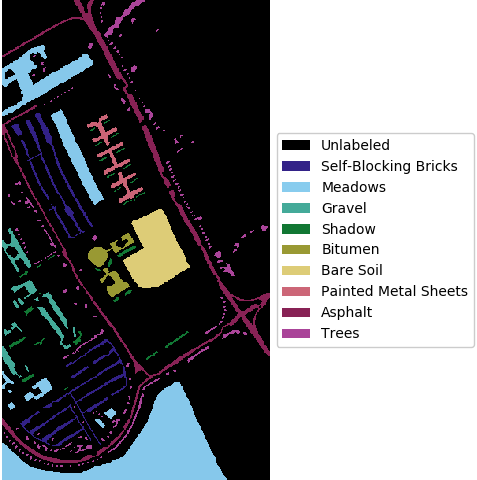}
  \label{fig:pavia_truth}}
  \subfigure[Indian Pines]{%
  \includegraphics[height=3cm]{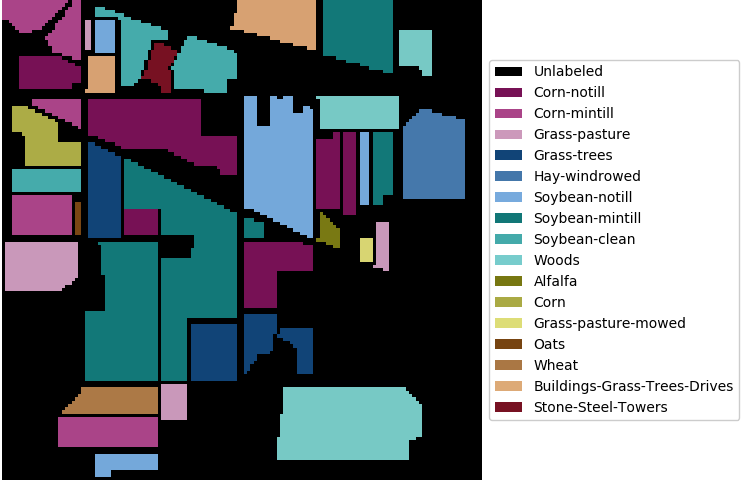}
  \label{fig:ip_truth}} 
  \caption{Ground truth maps for Indian Pines and Pavia University datasets.}
  \label{fig:datasets}
\end{figure}

\subsection{Low-Shot Learning Experiment}
In this experiment, we probe how each component of EarthMapper's pipeline affects performance with only a small quantity of training data (i.e., low-shot learning).  Thirty independent random trials are performed to calculate the mean and the standard deviation of the performance metrics--overall accuracy (OA), average accuracy (AA), and kappa coefficient. In each experimental trial, a training set and a validation set consisting of 15 and 35 randomly selected pixels belonging to each class are created. The pipeline is trained on these pixels and is evaluated on the remaining pixels in the image. We also measure the performance of the pipeline with and without SMCAE feature extraction and CRF-based post-processing. 

Table~\ref{table:results1} shows the classification results for our EarthMapper pipeline on the Indian Pines and Pavia University datasets.  In every case, the additional post-processing with the CRF improves classification performance, but the performance increase is less pronounced when we have a better (i.e., deeper) feature representation of the data. 

\begin{table*}[t]
\centering \footnotesize
\caption{Experimental results on Pavia University and Indian Pines HSI datasets, as the mean and standard deviation over 30 random trials, using 15 training/35 validation samples per class each trial.}
\begin{tabular}{@{}r|rrr|rrr@{}}\toprule
 & \multicolumn{3}{c|}{\textbf{Pavia University}} & \multicolumn{3}{c}{\textbf{Indian Pines}} \\
 &  \textbf{OA} & \textbf{AA} & {$\mathbf \kappa$} & \textbf{OA} & \textbf{AA} & {$\mathbf\kappa$} \\ \midrule
          \textbf{SS-MLP} & $78.8\pm3.2$ & $82.4\pm1.6$ & $0.727\pm0.037$ & $50.6\pm2.3$ & $65.9\pm1.9$  &$0.451\pm0.023$ \\
      \textbf{SS-MLP+CRF} & $82.5\pm4.0$ & $84.6\pm1.7$ & $0.773\pm0.047$ &$68.8\pm5.5$  & $83.3\pm2.9$ & $0.653\pm0.058$\\
    \textbf{SMCAE+SS-MLP} & $90.4\pm2.3$ & $92.1\pm1.2$ & $0.875\pm0.028$ & $84.3\pm1.6$ & $91.7\pm0.9$  & $0.823\pm0.018$\\
\textbf{SMCAE+SS-MLP+CRF} & \textbf{92.5} $\pm$ \textbf{1.8} & \textbf{92.6} $\pm$ \textbf{1.1} & \textbf{0.901} $\pm$ \textbf{0.023} & \textbf{88.3} $\pm$ \textbf{2.4} & \textbf{93.9} $\pm$ \textbf{1.1} &  \textbf{0.868} $\pm$ \textbf{0.027} \\
\bottomrule 
\end{tabular}
\label{table:results1}
\end{table*}

The combination of SMCAE feature extraction and CRF post-processing yields the best performance.  The performance increase from using SMCAE features \textbf{or} a CRF is statistically significant (paired $t$-test, $P < 0.01$) compared to when neither is used.  The performance increase from using SMCAE features \textbf{and} a CRF is statistically significant (paired $t$-test, $P < 0.01$) compared to using only one of the two.

\subsection{GRSS Data and Evaluation Server Results}
We also evaluated EarthMapper on the versions of the Indian Pines and Pavia University datasets provided by the IEEE GRSS Data and Algorithm Standard Evaluation (DASE) website.  Although these versions have more training samples, they are harder because the training samples are spatially co-located rather than randomly sampled throughout the scene, so the model sees less within-class variability.  We cross-validate SS-MLP using a 90\%/10\% training/validation (per-class) split.  Table~\ref{table:dase} compares EarthMapper's results  against the prior state-of-the art SuSA framework, which used a mean-pooled SMCAE feature response and the SS-MLP classifier~\cite{kemker2018low}.  The current EarthMapper configuration does not use a mean-pooling operation and uses a CRF for post-processing, since  mean-pooling can cross-contaminate different object classes for a given pixel.  The CRF worked very well with high-resolution HSI (i.e., Pavia University), but there was still a small performance increase for Indian Pines as well.

\begin{table}[t]
\centering \footnotesize
\caption{Experimental results for the IEEE GRSS versions of the Indian Pines and Pavia University datasets. }
\begin{tabular}{@{}r|rr@{}}\toprule
 & \textbf{Pavia University} & \textbf{Indian Pines}\\
 \midrule
 \textbf{SuSA}~\cite{kemker2018low}\\
 \textbf{OA} & 81.86 & 91.32\\
 \textbf{AA} & 74.09 & 81.17 \\
 {$\mathbf \kappa$} & 0.78 & \textbf{0.90}\\[1em]
 \textbf{EarthMapper}\\
 \textbf{OA}& \textbf{85.34}  & \textbf{91.59}\\
 \textbf{AA}& \textbf{ 78.93} & \textbf{82.70}\\
 {$ \mathbf \kappa$}  & \textbf{ 0.82}  & \textbf{0.90}\\
\bottomrule 
\end{tabular}
\label{table:dase}
\end{table}

\begin{figure}[t!]
  \centering
   \subfigure[Pavia University]{%
  \includegraphics[height=3cm]{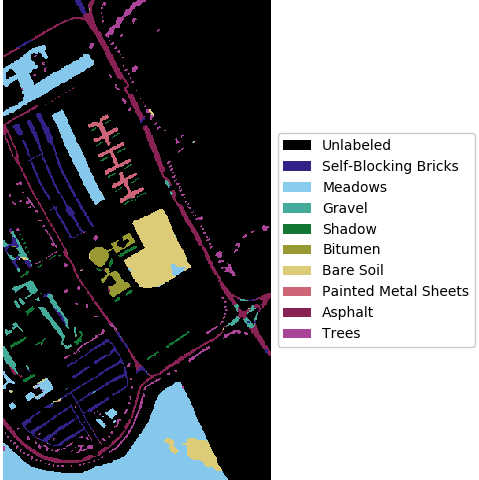}
  \label{fig:class_map_pavia}} 
  \subfigure[Indian Pines]{%
  \includegraphics[height=3cm]{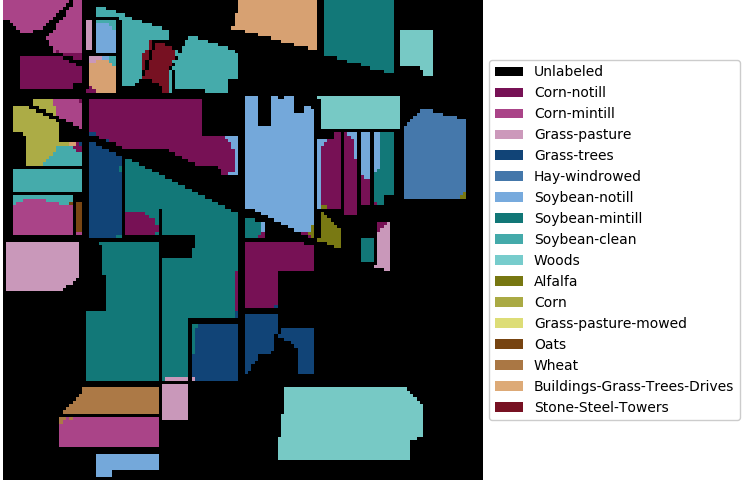}
  \label{fig:class_map_ip}}
  \caption{Classification maps for EarthMapper on the Indian Pines and Pavia University datasets from the Data and Algorithm Standard Evaluation website.}
  \label{fig:state_of_art}
\end{figure}

\section{Conclusion}

We have presented EarthMapper, a modular framework for the semantic segmentation of remote sensing imagery.  We demonstrate that incorporating deep, spatial-spectral feature extraction and UGM post-processing in the pipleline improves classification performance.  EarthMapper  yielded state-of-the-art performance for the two HSI benchmarks hosted on the GRSS DASE website.  EarthMapper is publicly available and can easily be adapted to incorporate new components. 

{\small
\bibliographystyle{ieee}
\bibliography{egbib}

\begin{thebibliography}{10}\itemsep=-1pt

\bibitem{alam2016crf}
F.~I. Alam, J.~Zhou, A.~W.-C. Liew, and X.~Jia.
\newblock {CRF learning with CNN features for hyperspectral image
  segmentation}.
\newblock In {\em IGARSS}, pages 6890--6893. IEEE, 2016.

\bibitem{alexandre2018interimage}
G.~Alexandre, O.~Pedro~da Costa, C.~Moutinho Duque~de Pinho, and R.~Feitosa.
\newblock Interimage: An open source platform for automatic image
  interpretation.
\newblock 03 2018.

\bibitem{spectral}
T.~Boggs.
\newblock Spectral python.
\newblock http://www.spectralpython.net, 2014.

\bibitem{boykov2001fast}
Y.~Boykov, O.~Veksler, and R.~Zabih.
\newblock Fast approximate energy minimization via graph cuts.
\newblock {\em IEEE Transactions on pattern analysis and machine intelligence},
  23(11):1222--1239, 2001.

\bibitem{Bunting:2014:RSG:2560972.2561467}
P.~Bunting, D.~Clewley, R.~M. Lucas, and S.~Gillingham.
\newblock The remote sensing and {GIS} software library.
\newblock {\em Comput. Geosci.}, 62:216--226, 2014.

\bibitem{cao2018hyperspectral}
X.~Cao, F.~Zhou, L.~Xu, D.~Meng, Z.~Xu, and J.~Paisley.
\newblock Hyperspectral image classification with markov random fields and a
  convolutional neural network.
\newblock {\em IEEE Transactions on Image Processing}, 2018.

\bibitem{dozat2016incorporating}
T.~Dozat.
\newblock Incorporating nesterov momentum into adam.
\newblock 2016.

\bibitem{pascal-voc-2012}
M.~Everingham, L.~Van~Gool, C.~K.~I. Williams, J.~Winn, and A.~Zisserman.
\newblock The {PASCAL} {V}isual {O}bject {C}lasses {C}hallenge 2012
  {(VOC2012)}.

\bibitem{flanders2003preliminary}
D.~Flanders, M.~Hall-Beyer, and J.~Pereverzoff.
\newblock Preliminary evaluation of ecognition object-based software for cut
  block delineation and feature extraction.
\newblock {\em Canadian Journal of Remote Sensing}, 29(4):441--452, 2003.

\bibitem{GDAL}
{GDAL Development Team}.
\newblock {\em GDAL - Geospatial Data Abstraction Library}.
\newblock Open Source Geospatial Foundation, 2018.

\bibitem{gewali2018tutorial}
U.~B. Gewali and S.~T. Monteiro.
\newblock A tutorial on modeling and inference in undirected graphical models
  for hyperspectral image analysis.
\newblock {\em arXiv preprint arXiv:1801.08268}, 2018.

\bibitem{Grizonnet2017}
M.~Grizonnet, J.~Michel, V.~Poughon, J.~Inglada, M.~Savinaud, and R.~Cresson.
\newblock {Orfeo ToolBox}: open source processing of remote sensing images.
\newblock {\em Open Geospatial Data, Software and Standards}, 2(1):15, Jun
  2017.

\bibitem{kemker2017self}
R.~Kemker and C.~Kanan.
\newblock Self-taught feature learning for hyperspectral image classification.
\newblock {\em IEEE TGRS}, 55(5):2693--2705, 2017.

\bibitem{kemker2018low}
R.~Kemker, R.~Luu, and C.~Kanan.
\newblock Low-shot learning for the semantic segmentation of remote sensing
  imagery.
\newblock {\em In review at IEEE TGRS}, 2018.

\bibitem{koller2009probabilistic}
D.~Koller and N.~Friedman.
\newblock {\em Probabilistic graphical models: principles and techniques}.
\newblock MIT press, 2009.

\bibitem{krahenbuhl2011efficient}
P.~Kr{\"a}henb{\"u}hl and V.~Koltun.
\newblock Efficient inference in fully connected {CRFs} with {Gaussian} edge
  potentials.
\newblock In {\em Advances in neural information processing systems}, pages
  109--117, 2011.

\bibitem{kussul2017deep}
N.~Kussul, M.~Lavreniuk, S.~Skakun, and A.~Shelestov.
\newblock Deep learning classification of land cover and crop types using
  remote sensing data.
\newblock {\em IEEE GRSL}, 14(5):778--782, May 2017.

\bibitem{lin2013spectral}
Z.~Lin, Y.~Chen, X.~Zhao, and G.~Wang.
\newblock Spectral-spatial classification of hyperspectral image using
  autoencoders.
\newblock In {\em Information, Communications and Signal Processing}, pages
  1--5. IEEE, 2013.

\bibitem{liu2015hyperspectral}
Y.~Liu, G.~Cao, Q.~Sun, and M.~Siegel.
\newblock Hyperspectral classification via deep networks and superpixel
  segmentation.
\newblock {\em Remote Sensing}, 36(13):3459--3482, 2015.

\bibitem{long2015fully}
J.~Long, E.~Shelhamer, and T.~Darrell.
\newblock Fully convolutional networks for semantic segmentation.
\newblock In {\em Proceedings of the IEEE conference on computer vision and
  pattern recognition}, pages 3431--3440, 2015.

\bibitem{ma2015hyperspectral}
X.~Ma, J.~Geng, and H.~Wang.
\newblock Hyperspectral image classification via contextual deep learning.
\newblock {\em EURASIP Journal on Image and Video Processing}, 2015(1):1--12,
  2015.

\bibitem{tong2018deep}
D.~H.~T. Minh, D.~Ienco, R.~Gaetano, N.~Lalande, E.~Ndikumana, F.~Osman, and
  P.~Maurel.
\newblock Deep recurrent neural networks for winter vegetation quality mapping
  via multitemporal sar sentinel-1.
\newblock {\em IEEE GRSL}, 15(3):464--468, March 2018.

\bibitem{tao2015unsupervised}
C.~Tao, H.~Pan, Y.~Li, and Z.~Zou.
\newblock Unsupervised spectral--spatial feature learning with stacked sparse
  autoencoder for hyperspectral imagery classification.
\newblock {\em IEEE GRSL}, 12(12):2438--2442, 2015.

\bibitem{tarabalka2010svm}
Y.~Tarabalka, M.~Fauvel, J.~Chanussot, and J.~A. Benediktsson.
\newblock Svm-and mrf-based method for accurate classification of hyperspectral
  images.
\newblock {\em IEEE GRSL}, 7(4):736--740, 2010.

\bibitem{valpola2015neural}
H.~Valpola.
\newblock From neural {PCA} to deep unsupervised learning.
\newblock {\em Advances in Independent Component Analysis and Learning
  Machines}, pages 143--171, 2015.

\bibitem{xia2015spectral}
J.~Xia, J.~Chanussot, P.~Du, and X.~He.
\newblock Spectral--spatial classification for hyperspectral data using
  rotation forests with local feature extraction and markov random fields.
\newblock {\em IEEE TGRS}, 53(5):2532--2546, 2015.

\bibitem{zhan2018semisupervised}
Y.~Zhan, D.~Hu, Y.~Wang, and X.~Yu.
\newblock Semisupervised hyperspectral image classification based on generative
  adversarial networks.
\newblock {\em IEEE GRSL}, 15(2):212--216, Feb 2018.

\bibitem{zhao2015combining}
W.~Zhao, Z.~Guo, J.~Yue, X.~Zhang, and L.~Luo.
\newblock On combining multiscale deep learning features for the classification
  of hyperspectral remote sensing imagery.
\newblock {\em Remote Sensing}, 36(13):3368--3379, 2015.

\bibitem{zhong2011modeling}
P.~Zhong and R.~Wang.
\newblock Modeling and classifying hyperspectral imagery by {CRFs} with sparse
  higher order potentials.
\newblock {\em IEEE TGRS}, 49(2):688--705, 2011.

\end{thebibliography}
}

\end{document}